\documentclass[letterpaper,10pt,conference]{ieeeconf}

\IEEEoverridecommandlockouts
\usepackage{times}
\usepackage{graphicx}
\usepackage{hyperref}
\usepackage{multicol}
\usepackage{amsmath}
\usepackage{bm}
\usepackage{capt-of}
\usepackage{caption}
\usepackage{amssymb}
\usepackage{booktabs}
\usepackage{graphicx}
\usepackage{bbm}
\usepackage{multirow}
\usepackage{siunitx}
\usepackage{wrapfig}
\usepackage{xspace}
\usepackage{xcolor}
\usepackage{stmaryrd}
\usepackage{trimclip}
\usepackage{hyperref}

\makeatletter
\DeclareRobustCommand{\shortto}{%
  \mathrel{\mathpalette\short@to\relax}%
}

\newcommand{\short@to}[2]{%
  \mkern2mu
  \clipbox{{.3\width} 0 0 0}{$\m@th#1\vphantom{+}{\shortrightarrow}$}%
  }
\makeatother

\definecolor{pmcolor}{RGB}{70,70,70}
\newcommand{\pmcolor}[1]{\textcolor{pmcolor}{#1}}
\newcommand{\pms}[2]{#1{\tiny{{\pmcolor{{$\pm$#2}}}}}}

\definecolor{red1}{RGB}{200,0,0}

\newcommand{\secv}{\vspace{-0.1em}}
\newcommand{\ssecv}{\vspace{-0.1em}}

\definecolor{citecolor}{HTML}{0071bc}
\hypersetup{
    colorlinks=true,
    citecolor=citecolor,
    filecolor=citecolor,
    linkcolor=citecolor,
    urlcolor=citecolor
}

\title{\LARGE \bf
\resizebox{\linewidth}{!}{From Simple to Complex Skills: The Case of In-Hand Object Reorientation}
}

\author{
Haozhi Qi$^{1,2}$, Brent Yi$^{1}$, Mike Lambeta$^{2}$, Yi Ma$^{1}$, Roberto Calandra$^{3,4}$ and Jitendra Malik$^{1,2}$\vspace{0.06in}\\
$^1$UC Berkeley \, $^2$FAIR at Meta \, $^3$TU Dresden \, $^4$Centre for Tactile Internet with Human-in-the-Loop\vspace{0.08in}\\
\href{https://dexhier.github.io}{\textcolor{citecolor}{https://dexhier.github.io}\xspace}
}

\begin{document}

\twocolumn[{%
    \renewcommand\twocolumn[1][]{#1}%
    \maketitle
    \centering
    \vspace{-1em}
    \includegraphics[width=\linewidth]{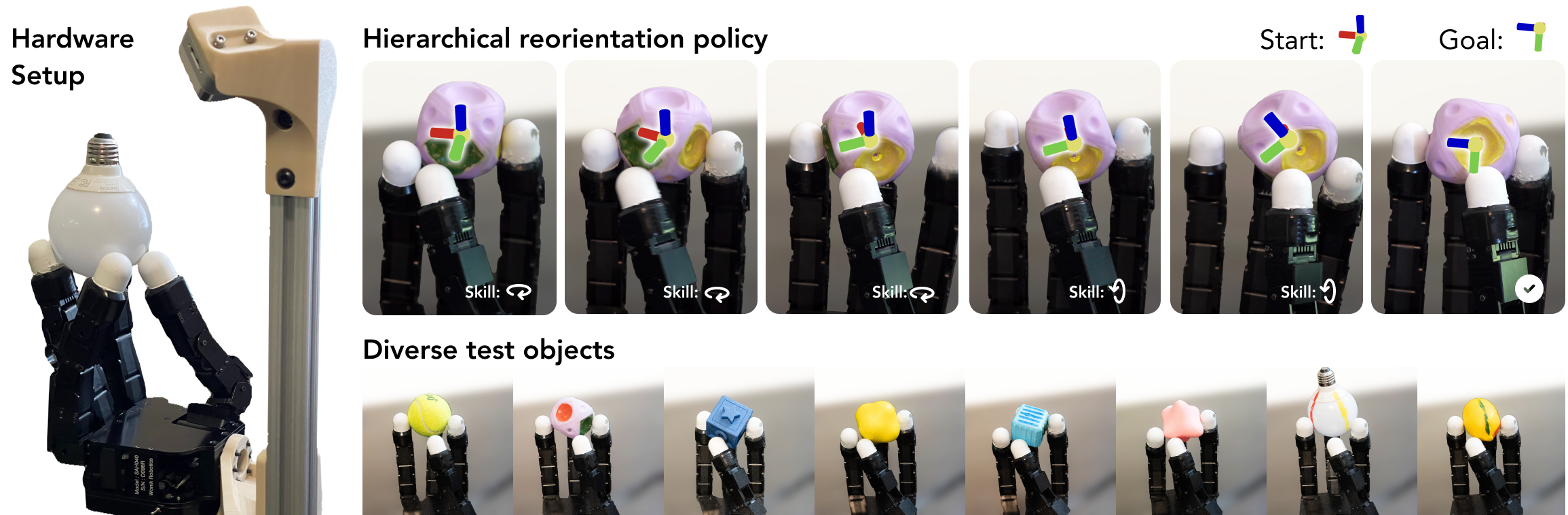}
    \captionof{figure}{\small
        \textbf{Left:} Hardware Setup. We use a multi-fingered robot hand with an RGB-D camera for our system. \textbf{Right:} We learn a hierarchical policy for in-hand object reorientation by reusing pre-trained skills (object rotation along single axes). It can manipulate diverse objects with symmetries and with different physical properties.
    }
    \label{fig:teaser}
    \vspace{4pt}
}]

\thispagestyle{empty}
\pagestyle{empty}

\begin{abstract}
Learning policies in simulation and transferring them to the real world has become a promising approach in dexterous manipulation. However, bridging the sim-to-real gap for each new task requires substantial human effort, such as careful reward engineering, hyperparameter tuning, and system identification.
In this work, we present a system that leverages low-level skills to address these challenges for more complex tasks.
Specifically, we introduce a hierarchical policy for in-hand object reorientation based on previously acquired rotation skills. This hierarchical policy learns to select which low-level skill to execute based on feedback from both the environment and the low-level skill policies themselves. Compared to learning from scratch, the hierarchical policy is more robust to out-of-distribution changes and transfers easily from simulation to real-world environments.
Additionally, we propose a generalizable object pose estimator that uses proprioceptive information, low-level skill predictions, and control errors as inputs to estimate the object's pose over time. We demonstrate that our system can reorient objects, including symmetrical and textureless ones, to a desired pose.
\end{abstract}
\vspace{-0.3em}

\secv
\section{Introduction}
\secv

Dexterous in-hand manipulation has made significant progress in recent years~\cite{openai2018learning, openai2019solving, qi2022hand}. A promising approach is to train a policy in simulation using reinforcement learning and then transfer it to the real world~\cite{chen2023visual, handa2023dextreme, qi2023general}. These policies are empirically generalizable and robust due to the diverse data available in simulation. However, they are typically trained from scratch for each task, requiring careful tuning of the reward function and its coefficients. This process demands substantial human effort and poses challenges for scalability.

In contrast, humans acquire new skills by building upon existing ones~\cite{diedrichsen2015motor}. Consider a beginner in tennis attempting their first serve: tossing a ball into the air, swinging their racket, and directing the serve. Each of these sub-skills is not practiced specifically within the context of tennis, but the individual can draw from past experiences with other balls or rackets. While the initial execution may be clumsy, it improves with practice.

Motivated by how humans acquire new skills, we argue that learning robot behavior for new tasks should leverage existing pre-trained skills. In machine learning, the use of pre-trained models has driven significant progress in computer vision~\cite{girshick2014rich, he2017mask} and natural language processing~\cite{devlin2018bert}, but its application to manipulation skill acquisition remains limited. To address this, we propose a hierarchical policy for in-hand object reorientation using pre-trained object rotation skills and demonstrate its effectiveness compared to training from scratch. We focus on this task because in-hand reorientation is a fundamental skill in daily life and exemplifies the complexity of dexterous manipulation.

The idea of exploiting hierarchies in robotics has also been widely studied in both task and motion planning~\cite{kaelbling2011hierarchical} and reinforcement learning~\cite{barto2003recent}. However, it often suffers because the low-level skill cannot provide enough feedback to the high-level policy, making it brittle if (inevitably) errors occur in the execution of the low-level skill. We tackle this problem by providing the high-level planner with feedback from the low-level skill and outputting a residual correction term to complement the low-level skills.

Specifically, we use the in-hand object rotation policies~\cite{qi2023general} as pre-trained skills and build a planner policy on top. The planner policy outputs two commands: 1) a rotation axis to guide the low-level rotation skills and 2) a residual action to complement the low-level actions. Such a design offers several advantages. First, it utilizes the structured low-level skills and reduces the exploration space, making training significantly more efficient and effective. Second, the low-level policy can predict an object representation and provide feedback to the high-level policy. This design enables the high-level policy to be aware of low-level skill reactions and correct errors by outputting residual actions. Additionally, since the low-level skills are transferable to the real world, the human effort required to bridge the gap is reduced.

Another challenge in building an in-hand object reorientation system is the need for a generalizable object state estimator. Although pose estimation has been extensively studied in computer vision~\cite{labbe2022megapose, wen2021bundletrack}, it suffers when encountering out-of-distribution objects and heavy occlusions, as seen in in-hand manipulation tasks. As a result, previous works bypass this problem by building state estimators using either object keypoints~\cite{handa2023dextreme} or point clouds~\cite{chen2023visual}. However, object keypoints must be manually designed for each object (e.g., the keypoints in~\cite{handa2023dextreme} refer to the vertices), and point clouds can't handle symmetrical objects well. For instance, even simple spheres are difficult to manage because point clouds look identical from any rotation angle.

To solve these problems, we propose a generalizable state estimator that takes a series of proprioceptive inputs and low-level skill feedback to predict the relative rotations over a given time interval. Our state estimator is inspired by~\cite{lennart2023estimator}, but unlike their approach, which requires training a separate policy for each object, our state estimator can generalize across different objects. We improve upon their method with two key changes: 1) using a modular design with a hierarchical policy, rather than learning the controller and state estimator together from scratch, and 2) incorporating feedback from low-level policies into the state estimator, allowing it to work with multiple objects. The estimator is trained entirely in simulation but is robust enough to handle novel objects in the real world.

We evaluate our system through comprehensive experiments. First, we demonstrate that our policy converges faster and achieves better performance compared to learning from scratch. Second, we show that we can learn a generalizable state estimator using the data generated in simulation and transfer it to the real world. We further study different design choices of the system through ablation experiments and verify the importance of residual actions and low-level skill feedback. Finally, we show that our learned policy naturally gains the benefits of a transferable low-level policy and can be easily transferred to the real world.

\secv
\section{Related Work}
\secv

\noindent\textbf{In-Hand Manipulation.} In-hand manipulation has been studied for decades~\cite{han1998dextrous, saut2007dexterous, rus1999hand, bai2014dexterous, mordatch2012contact, fearing1986implementing, teeple2022controlling, morgan2022complex, patidar2023hand, sieler2023dexterous} in classic robotics. More recently, learning-based methods have achieved significant progress~\cite{openai2018learning,qi2022hand,chen2023visual,khandate2023sampling,yang2024anyrotate,chen2024object,arunachalam2023holo,guzey2023dexterity}. One promising approach is to learn policies in simulation and transfer them to the real world (sim-to-real). This approach also demonstrates its flexibility for bimanual dexterous manipulation~\cite{lin2024twisting,huang2023dynamic}, dynamic tasks such as pen spinning~\cite{wang2024lessons}, and extensions to different modalities~\cite{yin2024learning, yin2023rotating,yuan2024robot,qin2023dexpoint}. These methods do not require an accurate dynamics model and can more easily leverage diverse available data. However, one of the major challenges that limits these methods from scaling up is the substantial human effort required to bridge the sim-to-real gap. In practice, this usually demands significant work for reward engineering, hyperparameter tuning, and domain randomization. Our work falls into the sim-to-real category but differs from others because we build a hierarchical policy for the in-hand manipulation system by reusing previously acquired skill policies instead of starting from scratch.

\noindent\textbf{Pose Estimation in Reorientation.} One key component of an in-hand object reorientation system is the pose estimation method. It has been extensively studied in both computer vision~\cite{wen2023bundlesdf} and robotics~\cite{labbe2022megapose, suresh2023neural}. However, pose estimation for in-hand manipulation is still far from being solved due to large occlusions and the requirements for efficiency and generalizability. As a result, previous work usually simplifies this problem either by assuming known object shapes~\cite{morgan2022complex, chen2023sequential} or by choosing to eschew general pose estimation. For example, Dextreme~\cite{handa2023dextreme} focuses only on a single cube and can thus estimate the pose using manually defined keypoints, but it cannot generalize to different objects. Visual Dexterity~\cite{chen2023visual} uses point clouds as a proxy for goal specification. However, this limits the flexibility of goal specification, as it cannot reorient a simple sphere because the point clouds look identical from all rotation angles. Recently, proprioceptive feedback has also been used to estimate pose~\cite{lennart2023estimator}. However, these methods train one policy to fit a single object and cannot generalize to different objects. Our work distinguishes itself by using generalizable low-level skills and utilizing feedback from low-level controllers, resulting in a generalizable pose estimator for multiple objects.


\noindent\textbf{Skill Hierarchy in Robotics.} The idea of hierarchy has a long history in classical robotics~\cite{kaelbling2011hierarchical,mason2001mechanics} and hierarchical reinforcement learning~\cite{pertsch2021accelerating, sutton1999between, bacon2017option}. The low-level controller can either be fine-tuned together with the high-level control or kept completely frozen~\cite{li2019sub, ranchod2015nonparametric, sharma2018directed, park2023hiql}. In the context of dexterous manipulation, Bhatt et al.~\cite{bhatt2022surprisingly} use manually designed action primitives to achieve open-loop dexterous manipulation. Khandate et al.~\cite{khandate2023dexterous} learn a switch between classic controllers and learning-based policies. Morgan et al.\cite{morgan2022complex} also decompose reorientation into several rotation sequences but do not use a learning-based method. HATO~\cite{lin2024learning} defines low-level grasping skills for easy teleoperation. In learning-based skill reuse, Gupta et al.~\cite{gupta2021reset} study reset-free reinforcement learning using multiple skills, but the task transitions are defined by the user. Sequential Dexterity~\cite{chen2023sequential} learns a transition feasibility function for sub-skill selection. In contrast, our approach does not require learning the transition explicitly; instead, the transition is implicitly encoded in the hierarchical policy.




\begin{figure*}[t]
\centering
\includegraphics[width=\linewidth]{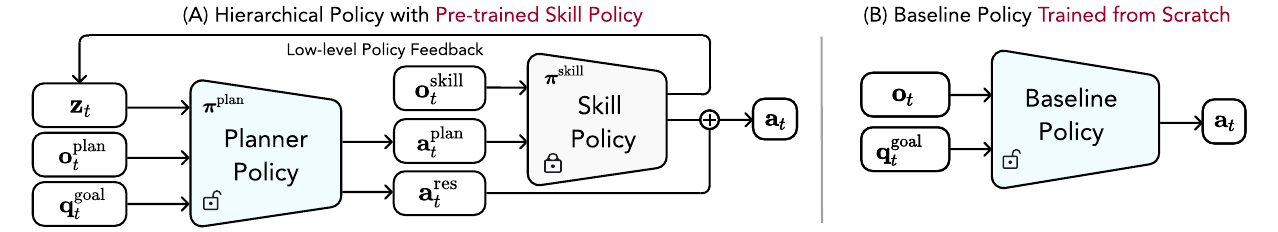}
\caption{\small \textbf{Comparison between our hierarchical policy and baseline policy.} Our planner policy takes goal orientation $\bm{q}_t^\text{goal}$, observation ${\bm{o}}^{\text{plan}}_t$, and feedback from low-level policies $\bm{z}_t$ as input and produces a one-hot skill vector $\bm{a}_t^\text{skill}$ along with a residual action $\bm{a}^{\text{res}}_t$, which is then used to control a low-level in-hand rotation skill.}
\vspace{-1.5em}
\label{fig:method}
\end{figure*}

\secv
\section{\small In-Hand Reorientation with Hierarchical Skills}
\secv

Our system overview is shown in Figure~\ref{fig:method} (A). It consists of two policies: the planner policy $\bm{\pi}^\text{plan}$ and the skill policy $\bm{\pi}^\text{skill}$. The planner policy $\bm{\pi}^\text{plan}$ takes an object's state, robot proprioception, feedback from low-level policies, and a goal orientation as input and outputs a rotation axis command $\bm{a}^\text{plan}_t$ and a residual action $\bm{a}^\text{res}_t$. This command is sent to the low-level policy $\bm{\pi}^\text{skill}$, which outputs the raw joint position targets $\bm{a}_t$ to the robot. To estimate the object's state in the real world, we additionally train a recursive estimator using feedback from sensory and low-level policies.

\ssecv
\subsection{Preliminary}
\ssecv

\noindent\textbf{Skill Policy.} Our skill policy is based on the in-hand object rotation policies in~\cite{qi2023general}. We select it because it demonstrates generalization in manipulating a diverse set of objects. We reimplement the framework and learn an axis-conditioned in-hand object rotation policy for a given axis $\bm{k}$.


Formally, the skill policy is defined as $\bm{a}_t^{\text{skill}}, \bm{z}_t = \bm{\pi}^\text{skill}(\bm{o}_t^{\text{skill}},\bm{k}_t)$ where $\bm{o}_t = [\bm{\theta}_{t-T:t}, \bm{a}^{\text{skill}}_{t-T-1:t-1}, \bm{d}_{t-T:t}]$. Among the observations, $\bm{\theta}_t \in \mathbb{R}^{16}$ represents the robot's joint positions, $\bm{a}^{\text{skill}}_t \in \mathbb{R}^{16}$ represents the commanded joint targets, and $\bm{d}_t \in \mathbb{R}^{32}$ represents the embedding of the depth image output by a lightweight convolutional neural network. We use $T=30$ in our experiments. The temporal sequence $\bm{o}_t$ is fed into a transformer, which outputs a single vector as the representation. The skill policy also outputs $\bm{z}_t$, which estimates the object's physical properties and shapes, represented by a feature vector. This vector serves as feedback from the low-level policy.

\noindent\textbf{Object State.} We define the object state space as $\bm{s}_t = \left[\bm{p}_t, \bm{q}_t\right]$, where $\bm{p}_t \in \mathbb{R}^{3}$ denotes the object's 3D position, and $\bm{q}_t \in \mathbf{S}^3$ denotes the object's orientation, represented as a unit quaternion. We define the relative pose as $\Delta(\bm{q}_{t_1}, \bm{q}_{t_2}) = \bm{q}_{t_2} \cdot \bar{\bm{q}}_{t_1}$, where $\bar{\bm{q}}$ denotes the conjugate of $\bm{q}$.

\ssecv
\subsection{Learning a Hierarchical Policy}
\label{sec:hier}
\ssecv

\begin{figure*}[t]
\includegraphics[width=\linewidth]{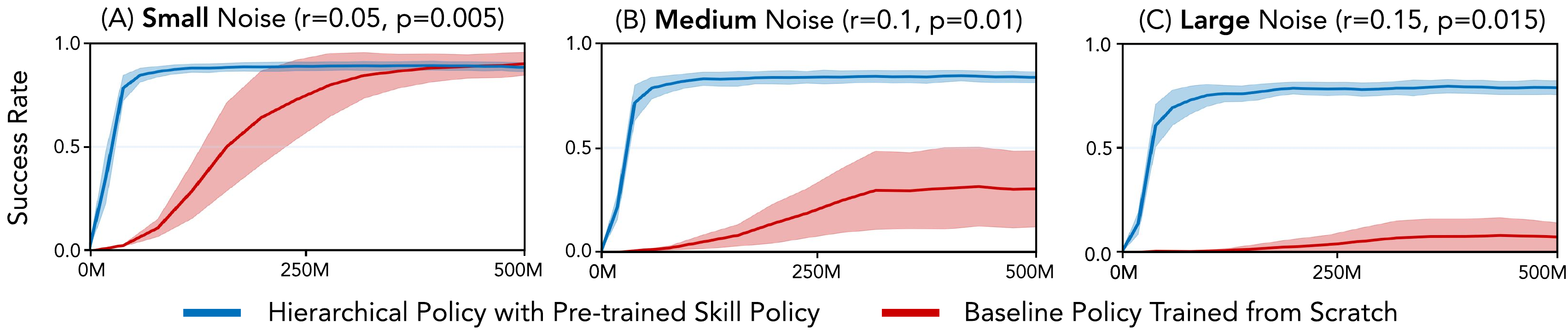}
\caption{\small \textbf{Policy training for different levels of object state noises.} We plot the training progress as the success rate with respect to agent steps. The shaded region represents one standard deviation from the mean success rate. In the easy case (A), both our method and the baseline achieve similar performance, but our method converges 8$\times$ faster than the baseline. As input noise is gradually increased, the baseline method becomes unstable and exhibits high variance across random seeds. In the case of large noise (C), the baseline method fails to converge, while our method maintains decent performance with low variance across seeds.}
\label{fig.compose}
\vspace{-1em}
\end{figure*}

\noindent\textbf{Observation and Action.} Our planner policy takes in object states, robot proprioception, feedback from low-level policies, and a desired orientation, and outputs the desired rotation axis $\bm{a}_t^{\text{plan}}$. To adapt to variance in object dynamics, we include a short horizon of paired robot states and control actions. Formally, we have $\bm{o}_t^{\text{plan}}=[\bm{s}_{t-5:t}, \bm{\zeta}_{t-5:t}, \bm{a}^\text{plan}_{t-6:t-1}]$ where $
\bm{a}_{t}^\text{plan}$ is the planner action in the previous timestep and $\bm{\zeta}_t = \Delta(\bm{q}_t, \bm{q}_t^\text{goal}) \in \mathbb{R}^4$ represents the relative transformation between object and goal orientation at timestep $t$. In addition, we augment the policy observation with the feedback from the low-level policy $\bm{z}_t$. Formally, we have $\bm{a}^{\text{plan}}_{t}= \bm{\pi}^\text{plan}(\bm{o}_{t}^{\text{plan}},\bm{q}_t^\text{goal}, \bm{z}_t)$. Note that, although the inputs to our policy do not explicitly contain the object's shape or physical property information, it is implicitly encoded in the low-level policy feedback $\bm{z}_t$.

In practice, we concatenate the inputs into a vector and pass it through the policy network. The policy network is a simple 3-layer MLP with ELU activation~\cite{clevert2015fast}. The network outputs a 7-dimensional categorical distribution, from which we sample a 7-dimensional one-hot action vector denoted as $\bm{a}_t^\text{plan}$. The dimensions correspond to one of the six canonical rotation axes ($\pm x$, $\pm y$, $\pm z$) and an additional \texttt{STOP} command. When inputting quaternions into the network, we convert them to 6D representations~\cite{zhou2019continuity}.

\noindent\textbf{Residual Actions.} Using a good set of low-level skills can accelerate exploration for new tasks. However, they cannot adapt to new tasks since they remain frozen during training. We augment the planner to output an additional residual action to complement the output of the skill policy. This design enables additional error correction from the planner policy. In summary, we have $[\bm{a}_{t}^\text{plan};\bm{a}_{t}^\text{res}] = \bm{\pi}^\text{plan}(\bm{o}_{t}^{\text{plan}},\bm{q}_t^\text{goal}, \bm{z}_t)$ and $\bm{a}_t = \bm{a}_t^{\text{res}} + \bm{a}_t^{\text{skill}}$ will be sent to the robot.

\noindent\textbf{Reward and Policy Optimization.} Our reward function is simple ($t$ omitted for simplicity) thanks for the robustness of pre-trained skill policy:
$r = 1 / (d(\bm{q}_t, {\bm{q}}^{\text{goal}}_t) + \epsilon) + \lambda_s \bm{1}({\rm Success})\,$,
where $\lambda_s$ are the coefficients for the reward terms. $1 / (d(\bm{q}_t, {\bm{q}}^{\text{goal}}_t) + \epsilon)$ is the rotational distance reward~\cite{openai2018learning,chen2023visual,handa2023dextreme}. $\bm{1}({\rm Success})$  is the success bonus used to encourage the planner to complete the task. Without this bonus, the policy learns to approach the goal but fails to finish it, as it aims to maximize the product of rotation reward and episode length. Compared to previous works~\cite{openai2018learning, chen2023visual, handa2023dextreme}, our reward function contains only two terms and is significantly easier to tune. This simplification is feasible because the low-level skills are already tuned to be transferable to the real world. Our central claim is that, when building policies for new tasks, we can avoid the tedious reward and hyperparameter tuning typically required for training from scratch.

In our design, the planner policy is trained using the ground-truth object states $\bm{q}_t$ provided by the simulator. We intentionally separate the perception and control components because this modularized design allows us to benefit from advancements in both reusable low-level skills and generalized perception models in the future.

\ssecv
\subsection{Generalizable State Estimator}
\ssecv
\label{sec:gse}

The policy described above takes noisy object state information from the simulator as input. To transfer it to the real world, a robust pose estimator is required for our system. Pose estimation has been extensively studied in computer vision; however, generalized pose estimation in uncontrolled environments remains an unsolved problem. Additionally, in-hand perception presents unique challenges. We propose a generalizable state estimator that takes a sequence of proprioceptive inputs and low-level skill feedback and outputs the relative rotations over a specified time interval. During deployment, the estimated relative transformations are integrated to determine if the desired pose is achieved.

\noindent\textbf{Recursive State Estimator.} Our state estimator is a neural network $\bm{\phi}$ that takes proprioception, action, control errors, low-level skill feedback, and previously estimated object state sequences as input and outputs the object state at the next timestep. We define the object's pose in the first frame as the canonical frame $\bm{q}_0$.

\noindent\textbf{State Estimation with Generalizable Low-level Skills.} Our approach differs from previous methods through the use of generalizable low-level skills and the feedback provided by these skills. Previous work requires training a separate policy for each object because it must simultaneously learn manipulation skills and object pose estimation~\cite{lennart2023estimator}. In contrast, since our low-level skills are generalizable, the pose estimator addresses a simpler task, enabling our policy to manipulate multiple objects. The skill policy not only facilitates the reorientation of diverse objects but also provides its own estimate $\bm{z}_t$ of the object's properties and shapes. For example, as shown in~\cite{qi2023general}, it encodes information about the object's geometry. This information is essential for achieving generalized object pose estimation.

We implement the state estimator using a transformer. We concatenate a temporal history of proprioception, action, control errors, and the predicted extrinsics as the input $\bm{f}_{t} = [\bm{q}_t, \bm{a}_{t-1},\bm{q}_t{\rm -}\bm{a}_{t-1},\bm{\hat{s}}_{t-1},\bm{z}_t].$
Then we feed a sequence of features $\bm{f}_{T} = \{\bm{f}_{t-k}, \dots,\bm{f}_{t-1},\bm{f}_{t}\}$ as input to the transformer. The transformer outputs $\bm{\hat{s}}_t$, which is used as input to our policy $\bm{\pi}^{\text{plan}}$.

To train this predictor, we rollout the planner policy $\bm{\pi}^{\text{plan}}$ with the \textit{predicted object state} in simulation. Meanwhile, we also save the ground-truth object state and construct a training set $\mathcal{B}=\{(\bm{f}_{T}^{(i)},\bm{s}_{t}^{(i)}, \hat{\bm s}_{t}^{(i)})\}_{i=1}^{N}.$ Then we optimize $\phi$ by minimizing the $\ell_2$ distance between $\bm{s}_t$ and $\hat{\bm{s}}_t$ using Adam~\cite{kingma2014adam}. Following~\cite{lennart2023estimator}, we reset the episode when the rotational distance between the predicted quaternion and the ground-truth exceeds 0.8 radians, or the predicted object location deviates by more than $3$ cm. This step is crucial for estimator training, as we find that the network does not converge without it.

\begin{figure*}[t]
\includegraphics[width=\linewidth]{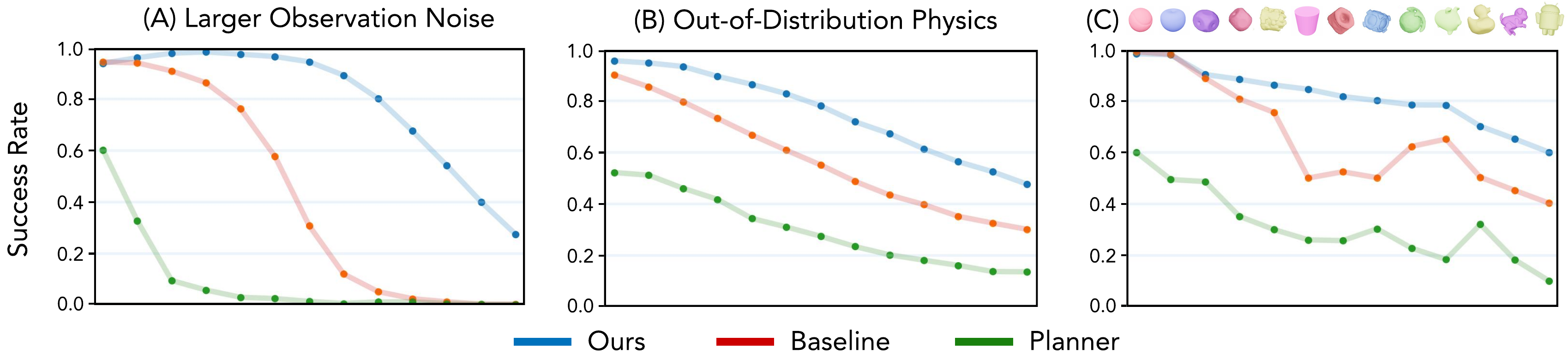}
\caption{\small \textbf{Robustness to different out-of-distribution scenarios.} We study the performance under larger orientation observation noise, physical randomizations, and shape variations. The perturbations become more challenging toward the right of each figure. Our policy and the baseline achieve similar success rates in easy cases but exhibit greater robustness in more challenging cases.}
\label{fig:robust}
\vspace{-0.2em}
\end{figure*}

\secv
\section{Experiments}
\secv

We first introduce the experiment setup in Section~\ref{sec.setup}. Then, we study the advantages of building a hierarchical policy by reusing in-hand object rotation skills compared to learning from scratch. We use the object state from the simulator as input and empirically analyze the training performance and robustness to out-of-distribution scenarios in Section~\ref{sec.s1_exp}. Next, we examine the performance when using the learned state estimator and the factors affecting the performance of the state estimator in Section~\ref{sec.s2_exp}. We also present ablation experiments on different design choices in Section~\ref{sec.ablation}, specifically showing the effectiveness of residual actions and low-level skill feedback. We demonstrate the sim-to-real results in the real world in Section~\ref{sec.real_exp}.

\ssecv
\subsection{Experiment Setup}
\label{sec.setup}
\ssecv

\noindent\textbf{Simulation Setup.} We use the IsaacGym simulator~\cite{makoviychuk2021isaac} to train our skill policy, planner policy, and state estimator. The simulation frequency is 120Hz, and the control frequency is 20Hz. We follow the standard setting~\cite{qi2022hand, khandate2023sampling} where the episode starts from a stable grasp sampled from a grasp set, and the target goal is randomly sampled from SO(3). For training, we use four different objects (cylinders, tennis balls, apples, and piggy banks) with randomized physics and sizes. Note that although the number of objects is not large, they still represent a class of different shapes. With randomized physical parameters, they provide enough variations to develop a robust policy~\cite{qi2022hand}.

\noindent\textbf{Fast Depth Rendering for In-Hand Objects.} One bottleneck in many previous in-hand manipulation works is the slow vision rendering speed. Previous studies have shown that this is due to IsaacGym not performing parallel rendering~\cite{gu2023maniskill2}. Therefore, we implemented our own pseudo-z-buffer depth rendering algorithm and increased the depth rendering speed from 800 FPS to 10,000 FPS.

\noindent\textbf{Hardware Setup.} We use an Allegro Hand from Wonik Robotics for our experiments. The Allegro Hand is a multi-fingered robot hand with four fingers and four degrees of freedom per finger. It is mounted on a specially designed hand frame connected to the camera (see Figure~\ref{fig:teaser}). The low-level policy receives vision input. We use MobileSAM~\cite{zhang2023faster} to detect the object in the first frame and Cutie~\cite{cheng2023putting} to track it over time. We use a Lambda workstation with an RTX4090 GPU, which is sufficiently fast for processing vision input and sending commands at 20 Hz.

\noindent\textbf{Baseline Methods.} We compare our approach with two baselines. First, we use a baseline policy trained from scratch, as shown in Figure~\ref{fig:method} (B). The baseline policy is also trained with PPO using similar reward and penalty terms. We comprehensively evaluate different observation choices, reward terms, and hyperparameters for the baseline and use the best set of hyperparameters for comparison. For observation, we use a short window of proprioception, depth encoding, robot actions, and the noisy object state as input. Second, we include a heuristic planner as a baseline. This planner receives noisy pose data and heuristically selects a rotation axis to send to the low-level policy.

\begin{table}[!t]
\centering
\setlength{\tabcolsep}{2pt}
\renewcommand{\arraystretch}{1.25}
\resizebox{\linewidth}{!}{%
\begin{tabular}{rrrrrrrrr}
\toprule
Method & Stage 1 $\shortto$ Stage 2 & Torque $\downarrow$ & Work $\downarrow$ & DofAcc $\downarrow$ & DofVel $\downarrow$ & LinVel 
$\downarrow$ & \\
\cmidrule(r){1-1}
\cmidrule(l){2-3}
\cmidrule(l){4-8}
Baseline & \pms{87.34}{12.13} $\shortto$ \pms{52.32}{4.89} &\pms{0.43}{0.24} & \pms{0.78}{0.55} & \pms{0.55}{0.23} & \pms{0.79}{0.30} & \pms{0.47}{0.10} \\
Ours & \pms{88.25}{1.28} $\shortto$ \pms{75.24}{1.27} & \pms{0.15}{0.01} & \pms{0.22}{0.01} & \pms{0.44}{0.02} & \pms{0.59}{0.02} & \pms{0.37}{0.01} \\
\bottomrule
\end{tabular}
}
\vspace{0.2em}
\captionof{table}{\small \textbf{Stage 2 and Policy smoothness evaluation.} We show the gap between stage 1 and stage 2 and evaluate several metrics, including energy metrics such as torque and work, robot smoothness metrics such as joint acceleration (DofAcc) and joint velocity (DofVel), and object stability (LinVel). Our method is significantly more stable than the baseline due to the use of smooth, transferable skills. Work and LinVel are scaled by 10x.}
\label{tab:energy}
\vspace{-1.5em}
\end{table}

\begin{figure*}[!t]
\includegraphics[width=\linewidth]{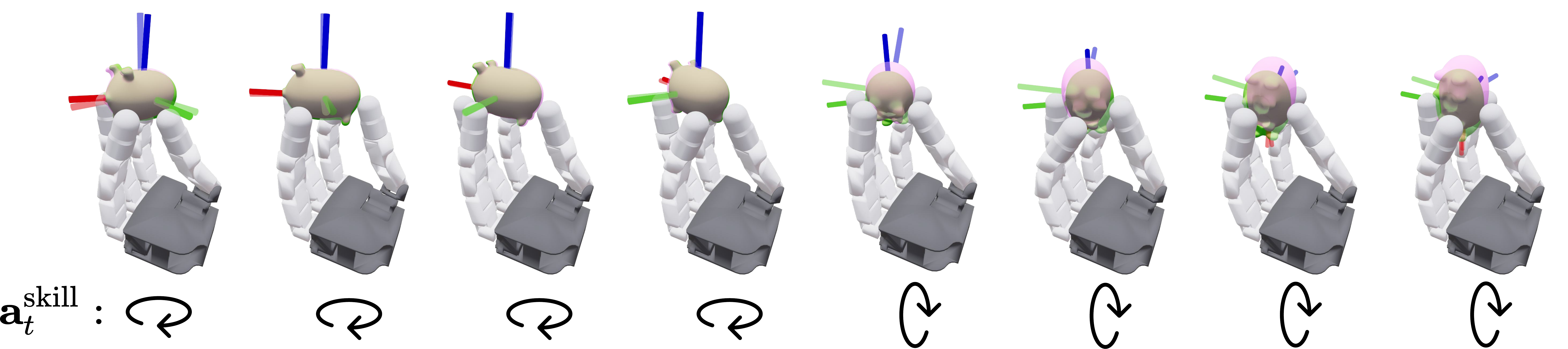}
\caption{\small \textbf{State estimation visualization across a rolled out trajectory.} We visualize poses from our learned estimator using the pink mesh and ground-truth poses using the green mesh. We observe generally robust estimation performance. Note that the sudden error in the middle is caused by slipping during the axis transition. Even in this challenging case, our state estimator can still predict a relatively accurate state and guide the planner to complete the task.}
\label{fig:dummy1}
\vspace{-1.3em}
\end{figure*}

\ssecv
\subsection{Policy Hierarchy with Pre-trained Skills}
\label{sec.s1_exp}
\ssecv

\noindent\textbf{Policy Learning Performance.} We first study the sample efficiency and training performance of our method compared to the baseline. The results are shown in Figure~\ref{fig.compose}. We examine three different levels of object state noise as input. Specifically, we add noise sampled from a normal distribution, with the standard deviation annotated in the figure title. $r=0.05$ represents a standard deviation of $0.05$ radians, and $p=0.005$ represents a standard deviation of $0.005$ meters.

In the small noise case (A), both our policy and the baseline achieve an 85\% success rate, but our policy converges 8$\times$ faster than the baseline. As we increase the noise level, we observe that the baseline policy becomes unstable and exhibits very high variance across different seeds (Figure~\ref{fig.compose} B), while our method remains stable. When we further increase the standard deviation of orientation noise from \SI{0.05}{\radian} to \SI{0.15}{\radian} and position noise from \SI{0.5}{\centi\meter} to \SI{1.5}{\centi\meter}, our policy remains stable, while the baseline policy fails to converge (Figure~\ref{fig.compose} C). The benefits of fast convergence and training stability come from the use of a pre-trained model, which provides a structured exploration space and avoids many meaningless random actions.

Notably, although the low-level skill policy has experienced additional samples compared to the baseline, simply increasing the training time of the baseline policy does not improve performance. We have tried to train the baseline policy with 20$\times$ more samples, but the conclusion remains the same. This experiment shows that the structural skill space of the low-level policy is more important than the number of samples it has seen during training.



\noindent\textbf{Out-of-Distribution Robustness.} We then study the out-of-distribution robustness of observation noise, physical randomizations, and object shapes for a \textit{trained model}. In this experiment, we use models trained under small noise from the previous section, as the baseline model tends to perform best in this training setting. The results are shown in Figure~\ref{fig:robust}. The general trend across all three evaluations indicates that, although the baseline performs similarly to our method in the easiest case (left data point in each plot), its performance drops rapidly as the test setting becomes more out-of-distribution.

Specifically, in the larger observation noise test, our policy maintains an 80\% success rate, while the baseline completely fails. When we increase the level of physical randomization, our method consistently outperforms the baseline. We also test shape generalization. For the tennis ball and apple, the baseline slightly outperforms our method. However, it fails to generalize well to other objects.

Another observation is that the heuristic planner performs significantly worse than both methods. Even in the easiest setting, it is noticeably inferior. This is because our low-level policy is not perfect, and a heuristic planner with predefined rules cannot account for this. As demonstrated in the ablation experiments, the design of the observation space for the planner plays a crucial role in performance.

\ssecv
\subsection{Generalizable State Estimation}
\label{sec.s2_exp}
\ssecv
In the previous section, the object states $\bm{p}_t$ and $\bm{q}_t$ are directly obtained from the simulator. However, to transfer the learned policy to the real world, we also need a general state estimator system for a diverse set of objects. In this section, we study task accuracy using our proposed state estimator and analyze how the policy's performance changes with predicted object states. Similar to the robustness test, we use the model trained with small noise and multiple objects.

\noindent\textbf{Comparison to Baseline.} We evaluate the policy's performance using the predicted object state as input. The results are shown in Table~\ref{tab:energy}. We compare the performance gap from stage 1 (using noisy object states) to stage 2 (using predicted states). Ideally, the performance gap would be small if the state estimation is accurate. We find that, although the baseline and our method achieve similar performance in stage 1, performance drops significantly in stage 2.

\noindent\textbf{Policy Smoothness.} To understand why the baseline policy experiences a larger performance drop, we quantitatively analyze various smoothness and energy metrics for our policy and the baseline. The energy metric is a critical factor for successful sim-to-real transfer~\cite{qi2022hand}. We measure energy metrics such as torque and work, robot smoothness metrics such as joint acceleration (DofAcc) and joint velocity (DofVel), and object stability (LinVel). Our results show that our method is significantly more stable than the baseline. As a result, object movement is smoother and easier to predict using sensory feedback. Note that we tune the baseline with all of these penalties and select the best-performing one without affecting the success metric.

Our policy achieves greater stability than the baseline due to the stability of the low-level policy. We argue that, as the task becomes more difficult, balancing task performance with the smoothness and stability required for sim-to-real transfer becomes increasingly challenging. Therefore, it is essential to use a structured low-level policy, and we hope our analysis provides solid evidence for this claim.

\noindent\textbf{Visualize Predicted Orientations.} We show a sequence of ground-truth object orientations, predicted orientations, and corresponding commanded actions in Figure~\ref{fig:dummy1}. This rollout is particularly challenging, as the object experiences slight slipping during the transition. Even in this scenario, our recursive estimator produces predictions that are accurate enough to guide the planner to complete the task. Additional video results are available in our supplementary material.

\begin{figure}[!t]
  \centering
  \includegraphics[width=\linewidth]{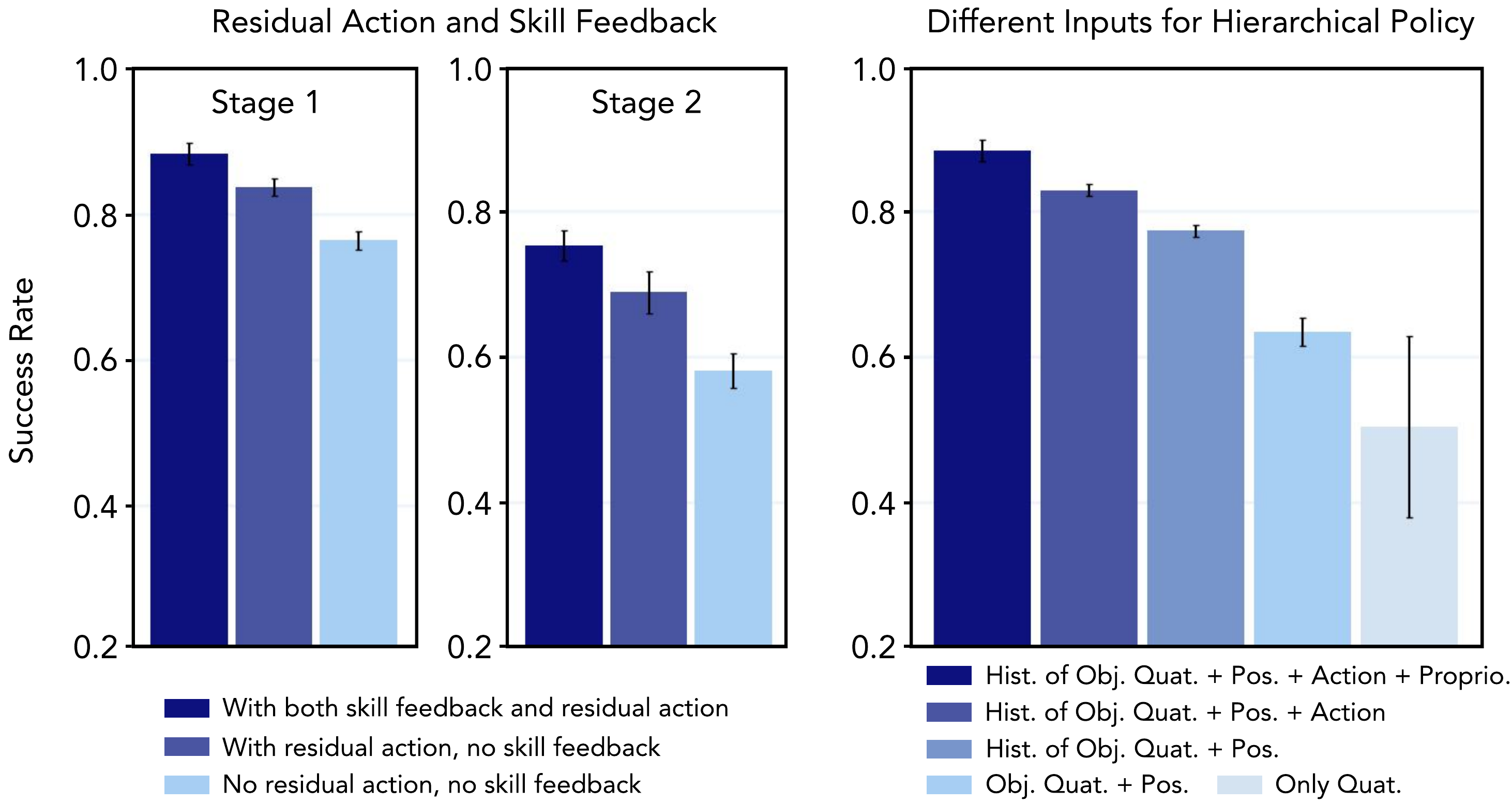}
  \caption{\small \textbf{Left:} Ablation experiments on residual actions and skill feedback from low-level skill policies. \textbf{Right:} The effect of different input formats to the planner policy.}
  \label{fig.ablation}
  \vspace{-1em}
\end{figure}

\begin{figure*}[!t]
\begin{minipage}[b]{0.40\linewidth}
\centering
\includegraphics[width=\linewidth]{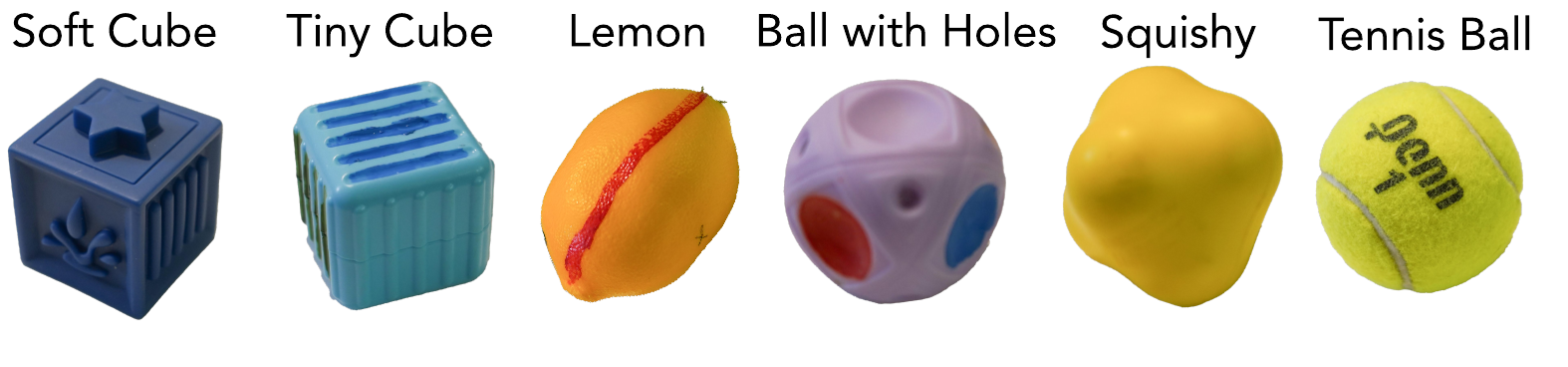}
\par\vspace{0.12em}
\end{minipage}
\hfill
\begin{minipage}[b]{0.58\linewidth}
\centering
\setlength{\tabcolsep}{2pt}
\renewcommand{\arraystretch}{1.55}
\resizebox{\linewidth}{!}{%
\begin{tabular}{lcccccccccccc}
\toprule
Method & \multicolumn{2}{c}{Soft Cube} & \multicolumn{2}{c}{Tiny Cube} & \multicolumn{2}{c}{Lemon} & \multicolumn{2}{c}{Ball w. Holes} & \multicolumn{2}{c}{Squishy} & \multicolumn{2}{c}{Tennis Ball} \\
& Single & Multi & Single & Multi & Single & Multi & Single & Multi & Single & Multi & Single & Multi \\
\cmidrule(r){1-1}
\cmidrule(r){2-3}
\cmidrule(r){4-5}
\cmidrule(r){6-7}
\cmidrule(r){8-9}
\cmidrule(r){10-11}
\cmidrule(r){12-13}
Ours & 73.3 & 60.0 & 70.0 & 37.5 & 86.7 & 85.0 & 76.7 & 65.0 & 80.0 & 75.0 & 93.3 & 90.0 \\
\bottomrule
\end{tabular}
} 
\par\vspace{0.3em}
\end{minipage}%
\caption{\small \textbf{Real-world performance for our policy.} We evaluate six different objects under two settings each and report the success rate. Our results show that our policy performs well in the real world for most of the objects.}
\label{fig.real}
\vspace{-1.3em}
\end{figure*}

\ssecv
\subsection{Ablation Experiments}
\label{sec.ablation}
\ssecv

In addition to the high-level design of reusing pre-trained skills, we also make several critical design choices within our architecture. In this section, all experiments are conducted under the small noise setting.

\noindent\textbf{Residual Actions and Low-Level Skill Feedback.} Our planner takes low-level skill feedback $\bm{z}_t$ as input and outputs a residual action on top of the skill policy's output. We show their effect in Figure~\ref{fig.ablation} (Left). Without these components, our method achieves 76.37\% accuracy in stage 1 (using noisy object states) and 58.12\% accuracy in stage 2 (using predicted object states), which is worse than training from scratch. With this basic design, adding residual actions improves performance to 83.63\% (stage 1) and 68.84\% (stage 2). This experiment shows that, even though the low-level policy is already tuned and achieves good performance on its tasks, performance can still be improved by adding small error corrections. When we further incorporate skill feedback into the policy, performance improves to 88.25\% and 75.24\%, indicating the importance of feedback. Note that, although our planner policy does not use vision information as input, the skill policy does take vision as input and predicts the object shape information encoded in $\bm{z}_t$.

\noindent\textbf{Policy Inputs.} Our planner incorporates several different types of observations (Section~\ref{sec:hier}), and we study the effect of each of them. The results are shown in Figure~\ref{fig.ablation} (Right). The most basic version (using only quaternion difference as input) achieves 50.37\% accuracy, which is similar to the heuristic planner baseline. Adding object position and including a history of observations both improve performance, as the policy can infer object dynamics from this information. Adding planner actions further improves performance, as the planner also accounts for the effects of its previous decisions. Finally, incorporating proprioception yields the best performance, highlighting the importance of being aware of the low-level robot state. This result underscores the significance of closed-loop feedback between the planner and the low-level skill policies.

\ssecv
\subsection{Real-World Experiments}
\label{sec.real_exp}
\ssecv
Finally, we transfer the learned policies to the real world. We test our policies on six objects (Figure~\ref{fig.real}, left), using proprioception and segmented depth as inputs. Note that there is no cube in our training set, so most of the real-world objects are out-of-distribution. We evaluate two settings: \textit{Single:} We set the target to rotate along one of the $x/y/z$-axes by $\pi/2$ or $\pi$ radians. This setting does not require switching between different skills but tests the accuracy of pose estimation. We evaluate 30 trials per object. \textit{Multi:} We set the target so that the policy needs to rotate over two axes by $\pi/2$. We evaluate 20 trials per object.

The results are shown in Figure~\ref{fig.real}. We find that our policy performs well in the real world for most of the objects. The most challenging object is the tiny cube, as it is small relative to the size of the hand, making it difficult to manipulate. The successful sim-to-real transfer is attributed to two factors: 1) the use of transferable low-level skills and 2) a well-designed planner structure, as demonstrated in the ablation experiments (Section~\ref{sec.ablation}). We also provide qualitative videos showcasing the reorientation in the real world in our supplementary material.

\secv
\section{Conclusions and Limitations}
\label{sec:conclusion}
\secv

In this work, we present a system for in-hand object reorientation by building a hierarchical policy with pre-trained low-level skills. To achieve robust and generalizable state estimation in the real world, we also learn a state estimator. We demonstrate successful deployment on multiple symmetric and textureless objects. Our work highlights the potential of moving away from robot policies trained from scratch: for a given task, we can significantly improve training efficiency, robustness, and generalizability by leveraging pre-trained models for lower-level skills.

\noindent\textbf{Limitations and Future Work.} Our approach relies on the effectiveness of generalizable low-level policies. Importantly, it requires that no slipping occurs between the finger and the object. This issue can be mitigated by incorporating tactile sensing into our framework. In our current setting, estimated pose errors accumulate over time. Integrating vision and touch for accurate and long-term pose tracking is a promising direction for future work.

\section*{Acknowledgment}
This research is supported as a BAIR Open Research Common Project with Meta. In their academic roles at UC Berkeley, Haozhi Qi and Jitendra Malik are supported in part by DARPA Transfer from Imprecise and Abstract Models to Autonomous Technologies (TIAMAT) program and ONR MURI N00014-21-1-2801. Brent Yi is supported by the NSF Graduate Research Fellowship Program under Grant DGE 2146752, and Haozhi Qi, Brent Yi, and Yi Ma are partially supported by ONR N00014-22-1-2102 and the InnoHK HKCRC grant. Roberto Calandra is supported by the German Research Foundation (DFG, Deutsche Forschungsgemeinschaft) as part of Germany’s Excellence Strategy – EXC 2050/1 – Project ID 390696704 – Cluster of Excellence “Centre for Tactile Internet with Human-in-the-Loop” (CeTI) of Technische Universität Dresden, and by the Bundesministerium für Bildung und Forschung (BMBF) and German Academic Exchange Service (DAAD) in project 57616814 (\href{https://secai.org/}{SECAI}, \href{https://secai.org/}{School of Embedded and Composite AI}). We thank Jiayuan Mao and Qiyang Li for their valuable feedback on related literature. We thank Xinru Yang for her help on real-world videos.

\bibliographystyle{IEEEtran}
\bibliography{IEEEabrv,references}

\end{document}